\documentclass[runningheads]{llncs}

\usepackage{eccv}

\usepackage[table]{xcolor}
\usepackage{microtype}
\usepackage{graphicx}
\usepackage{array}

\usepackage{amsmath}
\usepackage{amssymb}
\usepackage{mathtools}
\usepackage{pifont} 

\usepackage{caption, subcaption, multirow, overpic, textpos, booktabs}
\usepackage[pagebackref=false, breaklinks=true, letterpaper=true, colorlinks,
            citecolor=citecolor, linkcolor=linkcolor, bookmarks=false]{hyperref}
\definecolor{citecolor}{HTML}{0071BC}
\definecolor{linkcolor}{HTML}{ED1C24}

\newlength\savewidth\newcommand\shline{\noalign{\global\savewidth\arrayrulewidth
  \global\arrayrulewidth 1pt}\hline\noalign{\global\arrayrulewidth\savewidth}}
\newcommand{\tablestyle}[2]{\setlength{\tabcolsep}{#1}\renewcommand{\arraystretch}{#2}\centering\footnotesize}
\renewcommand{\paragraph}[1]{\vspace{1.25mm}\noindent\textbf{#1}}

\newcolumntype{x}[1]{>{\centering\arraybackslash}p{#1pt}}
\newcolumntype{y}[1]{>{\raggedright\arraybackslash}p{#1pt}}
\newcolumntype{z}[1]{>{\raggedleft\arraybackslash}p{#1pt}}

\definecolor{deemph}{gray}{0.6}

\definecolor{baselinecolor}{gray}{.9}
\newcommand{\baseline}[1]{\cellcolor{baselinecolor}{#1}}

\usepackage{marvosym}
\usepackage{tikzsymbols}

\usepackage{eccvabbrv}

\usepackage{graphicx}

\usepackage[accsupp]{axessibility}  

\usepackage{orcidlink}

\begin{document}

\title{DailyMAE: Towards Pretraining Masked Autoencoders in One Day} 

\titlerunning{DailyMAE}

\author{Jiantao Wu\inst{1}\thanks{jiantao.wu@surrey.ac.uk} ~~ Shentong Mo\inst{2} ~~
Sara Atito\inst{1} ~~ Zhenhua Feng\inst{1} ~~ Josef Kittler\inst{1} ~~ Muhammad Awais\inst{1}
}
\authorrunning{J.~Wu et al.}

\institute{University of Surrey \and
Carnegie Mellon University}

\maketitle

\vspace{-0.1cm}
\begin{abstract}
Recently, masked image modeling (MIM), an important self-supervised learning (SSL) method, has drawn attention for its effectiveness in learning data representation from unlabeled data. Numerous studies underscore the advantages of MIM, highlighting how models pretrained on extensive datasets can enhance the performance of downstream tasks. However, the high computational demands of pretraining pose significant challenges, particularly within academic environments, thereby impeding the SSL research progress. In this study, we propose efficient training recipes for MIM based SSL that focuses on mitigating data loading bottlenecks and employing progressive training techniques and other tricks to closely maintain pretraining performance. Our library enables the training of a MAE-Base/16 model on the ImageNet 1K dataset for 800 epochs within just 18 hours, using a single machine equipped with 8 A100 GPUs. By achieving speed gains of up to 5.8 times, this work not only demonstrates the feasibility of conducting high-efficiency SSL training but also paves the way for broader accessibility and promotes advancement in SSL research particularly for prototyping and initial testing of SSL ideas. The code is available in \href{https://github.com/erow/FastSSL}{FastSSL}

\end{abstract}

\section{Introduction}
\label{sec:intro}

Self-Supervised Learning (SSL) represents a transformative shift in machine learning, by harnessing unlabeled data for effective model training~\cite{oord_representation_2018,atito_sit_2021,atito_gmml_2022,he_masked_2022,grill_bootstrap_2020,mo2021spcl,mo2022pauc,mo2023representation}. Such a learning paradigm benefits from large-scale datasets to learn rich representations for few-shot learning~\cite{brown_language_2020} and transfer learning~\cite{chen_empirical_2021,he_momentum_2020}. 
Vast amount of unlabeled data on internet ignite the need to scale training of deep neural network models to large datasets. 
Currently, the success of SSL usually requires weeks of training on high performance clusters (HPC)~\cite{brown_language_2020,chen_generative_2020,devlin_bert_2019}.
For instance, iBOT~\cite{zhou_ibot_2022} is trained for 193 hours on 16 V100s for ViT-S/16. These calculations don't include testing of different hypothesis while developing a SSL framework which need to be tested on a decent scale of ImageNet-1K~\cite{russakovsky_imagenet_2015} with 1.2 millions samples for a considerable number of epochs. 
Therefore, efficient pretraining recipes are highly desired to accelerate the study of SSL algorithms, e.g., hyper-parameter tuning and fast validation for novel algorithms.
To reduce the training time, some researchers train their models on a subset of ImageNet-1K~\cite{russakovsky_imagenet_2015}, e.g., 10\% samples~\cite{atito_sit_2021}. However, there can be a performance gap when scaling the model to large datasets, that is, a model sophisticated in small datasets may not handle the diversity on complex problems.

A lot of effort has been made to improve the training efficiency through Mix-Precision~\cite{micikevicius_mixed_2018}, FlashAttention~\cite{dao_flashattention_2022}, efficient architectures~\cite{he_masked_2022,tan_efficientnetv2_2021} and so on. 
FFCV~\cite{leclerc_ffcv_2023} is proposed to remove the data loading bottleneck, which increases a throughput of  images/s. Based on it, FFCV-SSL~\cite{bordes_towards_2023} accelerates SimCLR from 32 hours to 8 hours on 8 V100s. 
Masked autoencoders (MAEs)~\cite{he_masked_2022}, by removing the masked tokens, accelerate training (by 3× or more) and achieve competitive performance. 
Despite the efficiency of MAEs, the original implementation still requires about 800 GPU hours (V100) to reproduce the results. 


\begin{figure}[t]
\begin{minipage}[t]{.4\linewidth}
    \centering
    \includegraphics[width=.9\linewidth]{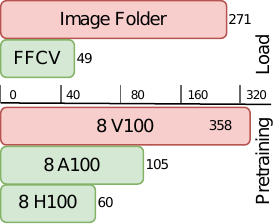}
    \caption{\textbf{Time consumption} for loading and training one epoch (1,281,167 images). The runtime of pretraining MAE-B/16 is measured without data loading.} \label{fig:runtime}
\end{minipage}
\hspace{.05\linewidth}%
\begin{minipage}[t]{.55\linewidth}
    \includegraphics[width=\linewidth]{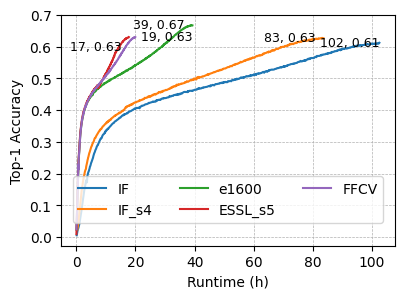}
    \caption{\textbf{Online prob} for pretraining mae-base/16 from scratch with respect to the training time (x-axis) on a single machine with 8 A100s. Each point denotes the total runtime and final accuracy. ``ESSL\_s4'' denotes our improved FFCV with dynamic resolution. }\label{fig:online_prob}
\end{minipage}

\end{figure}

In this work, we aim to optimize the training efficiency on a single machine. Although the hardware has been improved dramatically in recent years, e.g. the H100 released in 2022 is 3x faster than V100 released in 2017,  the data loading has become a major bottleneck for training a base model, as illustrated in Fig.~\ref{fig:runtime}. Yet, most of the machine learning libraries still use the Pytorch implementation, reading images from folders (IF), which significantly hinders the training process. To remove this bottleneck, we utilize FFCV~\cite{leclerc_ffcv_2023} for fast data loading. We further study the effects of maximum resolution and image quality for compression, which is crucial to the trade-off between storage, throughput, and performance. In addition, we propose `crop decode' to optimize FFCV for faster decoding and lower memory usage. Our improved FFCV, termed ESSL, is 27.6\% faster and saves 13.7\% memory compared to the original implementation. 

The training process can be further sped up by progressive training, where the image size is gradually increasing during training~\cite{tan_efficientnetv2_2021,howard_fastai_2018}. A lot of effort has been made to train models on smaller images. One of advantages is to reduce the number of tokens, leading to lower computational costs. Another advantage is the regularization effect of preventing overfitting~\cite{touvron_deit_2022,li_inverse_2023}. By gradually enhancing the data augmentation, the model can converge to a better minimum. However, most of the studies about progressive training are based on convolutional neural networks (CNNs) and focus on supervised learning. We revisit the application of progressive training with ViTs in mind for finetuning, because the image size is relative to the apparent size seen by a patch embedding. 
We further apply progressive training into pretraining. Surprisingly, a palindrome scheme for progressively decreasing and then increasing the image size during training, maintains the competitive performance and reduces training time. 
Our contributions can be summarized as
\begin{itemize}
    \item We propose a machine learning library for masked autoencoders, where our pretraining process of mae-base/16 is 5.8x faster than the official one.
    \item We introduce the `crop decode' to reduce the memory usage and accelerate the loading pipeline. We discuss the effects of compression with regard to the resolution and image quality.
    \item We propose a novel pretraining strategy utilizing a palindrome scheme. This approach surprisingly maintains competitive performance while achieving a 10.9\% reduction in training time.
\end{itemize}

\section{Related Work}

\paragraph{Data Loading Library.}
For a machine learning system with a single node, loading data samples to tensors is becoming the bottleneck of machine learning, due to the fast development of AI chips to speed up the computation, but only a few improvements being made for the data file handling. A general data loading pipeline consists of reading and processing. DALI~\cite{noauthor_nvidiadali_2024} is a high-performance alternative to the pytorch data loaders, which increases the throughput by processing data with GPUs. However, reading individual images from folders slows the pipeline due to random read for small files. WEBDATASET~\cite{noauthor_webdatasetwebdataset_nodate} optimizes the I/O rates by gathering samples into tar files. By optimizing the data loading and processing, Fast Forward Computer Vision (FFCV)~\cite{leclerc_ffcv_2023} eliminates data bottlenecks, combining techniques such as an efficient file storage format, caching, data pre-loading, asynchronous data transfer, and just-in-time compilation. Noticing that a typical pipeline decodes the whole image and then crops a part of it, in our work we introduce the \textit{crop decode} operation to avoid decoding the discarded parts and apply it within FFCV to accelerate the data loading.

\paragraph{Masked Image Modeling.}
Self-supervised learning is becoming a hot research topic in machine learning to address the limitations of supervised learning, which requires large amounts of labeled data~\cite{oord_representation_2018,grill_bootstrap_2020,chen_simple_2020,wu2022objectwise}. 
Masked Image Modeling (MIM) is one of the successful approaches that excel in learning visual representations from extensive datasets~\cite{atito_gmml_2022,he_masked_2022,xie_simmim_2022,wu2023masked,wu2023accuracy}. It takes inspiration from the successes in masked language models in NLP~\cite{devlin_bert_2019}, employing a similar strategy of masking and prediction to enable neural networks to grasp the intricate structure and context of visual data. MIM's efficacy in bolstering the downstream task performance, including image classification and semantic segmentation, is a testament to its ability to learn transferable and generalized features~\cite{atito_gmml_2022,bao_beit_2022,he_masked_2022}.

In the landscape of MIM, diverse masking strategies have been proposed. GMML~\cite{atito_gmml_2022}, for instance, integrates group masking and alien blocks to effectively disrupt and obscure input data for enhanced feature learning. BeiT~\cite{bao_beit_2022} employs mask tokens as placeholders for masked patches, offering a novel approach to handle occlusions in visual data. The masked autoencoder~\cite{he_masked_2022} applies an asymmetric structure, with an encoder dropping the masked patches, along with a lightweight decoder reconstructing the original images from the latent representations and mask tokens. Many improvements have been proposed to optimize the efficiency and effectiveness. For instance, Data2vec 2.0~\cite{baevski_efficient_2023} optimizes the data-efficiency of the objective  by learning rich contextualized targets.
Though there are other improvements~\cite{wu_tinyvit_2022,wang_closer_2023,zhou_ibot_2022} having higher performance or lower cost. Our work focuses on the original masked autoencoder~\cite{he_masked_2022} for the sake of simplicity. Our framework is easy to use and has the capacity to adopt these improvements in the future.

\paragraph{Progressive Training.}
A lower resolution, e.g., 192x192, benefits the training process. On the one hand, it significantly reduces the training time and memory usage, especially for transformers. On the other hand, it has a regularization effect, which enforces the model to learn a general sense of the images~\cite{touvron_deit_2022,cao_rethinking_2021}. 
Motivated by FixRes~\cite{touvron_fixing_2022}, which found that the discrepancy between the training set and the test set in the sense of working with different sizes of the objects causes performance drop on validation, we study the effect of the scaling ratio for cropping images. The inverse scaling law~\cite{li_inverse_2023} shows that only a large model gains improvements from fewer tokens.
Therefore, there is a trade-off between reducing the generalization error and the training error for the resolution.
Progressive resizing, gradually increasing the image size during training, is introduced for image classification~\cite{howard_fastai_2018,bordes_towards_2023,ridnik_solving_2022}, video summary~\cite{haopeng_progressive_2022}, and for vision~\cite{jerrish_deep_2023}. Such a simple technique not only accelerates the training process but also reduces the generalization error. EfficientNetV2~\cite{tan_efficientnetv2_2021} further improves the progressive training of merely increasing the image size by gradually adding stronger regularization.
However, the merits of the method of progressive training for pretraining have not been properly explored.

\section{Efficient Masked Autoencoder}

In this study, we develop efficient training strategies for masked autoencoders, focusing on the pretraining and finetuning stages. We primarily utilize the Vision Transformer base model (~85M)~\cite{dosovitskiy_image_2020} for training on ImageNet-1K. Although adopting other efficient architectures like Swin Transformers~\cite{liu_swin_2022} and Hierarchical models~\cite{huang_green_2022,ryali_hiera_2023}, or advanced models like Evolved MAE~\cite{feng_evolved_2023} and SdAE~\cite{avidan_sdae_2022}, could further reduce the training time, we chose the basic MAE for its simplicity and wide applicability. This decision was driven by two factors: 1) MAE's compatibility with various models requiring minimal modifications, and 2) our aim to establish a foundational library to facilitate future innovations, where a complex model might hinder novel developments.

Our study utilizes an enhanced FFCV~\cite{leclerc_ffcv_2023} to eliminate data loading delays and employs progressive training for gradually resizing images without compromising performance. These straightforward methods considerably speed up the learning process. THe comparison of training processes is depicted in Fig.~\ref{fig:online_prob}, showing significant acceleration in the pretraining stage. Table~\ref{tab:perf} demonstrates our efficient benchmarks. We differentiate between the official MAE pretrained and finetuned weights, denoted as \textbf{MAE} and \textbf{ft-MAE}\footnote{\href{https://github.com/facebookresearch/mae}{Official MAE implementation.}}, and our implementation's from-scratch models, referred to as \textbf{mae} and \textbf{ft-mae}. Our training recipes are summarized in Table~\ref{tab:recipe}.

\paragraph{Fintuning Recipe.}
Our finetuning recipe adopts the MAE finetuning settings but opts for a different data augmentation strategy. Instead of the commonly used RandAug~\cite{cubuk_randaugment_2020}, which automatically selects combinations of augmentation operations to apply to the training data, but falls short in terms of data shifts from compression, we employ the Three Augmentation (3 Aug)~\cite{touvron_deit_2022} procedure, with better resilience on compressed datasets. It is notable that the model produces different results on validation sets with different compression parameters. Thus, we standardize the validation set at the maximum resolution of 500 and quality of 100, denoted as Val(500\_100), enhancing the model evaluation consistency. This adjustment, along with progressive training, cuts the finetuning time from 21 to 18 hours on a setup with 8 V100 GPUs.

\paragraph{Pretraining Recipe.}
The masked autoencoder~\cite{he_masked_2022}, a popular self-supervised architecture, significantly speeds up training by omitting masked patches, achieving a 3x increase in speed. However, data loading becomes a bottleneck due to the rapid hardware advancements depicted in Fig.~\ref{fig:runtime}. Our approach enhances the efficiency by substituting the  data loader with an improved FFCV and a dynamic image resolution strategy. Unlike the traditional progressive learning method, which increases the resolution gradually, we find that decreasing and then increasing the resolution during training is much more beneficial for MAEs.
These improvements significantly reduce the pretraining time, diminishing it to 17 hours with progressive training, while maintaining the MAE high performance.

\begin{table}[t]

\begin{minipage}[t]{.52\linewidth}
    \centering
    \caption{\textbf{Training recipes} for finetuning and pretraining. Runtime is measured on eureka (single machine with 8 A100s).}\label{tab:recipe}
    \begin{tabular}{l|cc}
    Recipe&  finetune &pretrain\\
    \shline
    Batch size & 1024  &4096\\
    Optimizer & AdamW   &AdamW\\
    LR & $2^{-3}$  &2.4e-3\\
    LR schedule& cosine  &cosine\\
     Layer-wise lr decay& 0.65  &-\\
    Weight decay & 0.05  &0.05\\
    Warmup epochs & 5  &40\\
     epochs&100  &800\\
    Augmentation &  3 Aug~\cite{touvron_deit_2022} & Simple~\cite{he_masked_2022}\\
     \end{tabular}
\end{minipage}
\hspace{.02\linewidth} %
\begin{minipage}[t]{.45\linewidth}
\centering
\caption{\textbf{Benchmark} of our library. ``fixed'' denotes fixed resolution of 224. ``e1600'' denotes training 1600 epochs. }\label{tab:perf}
\begin{tabular}{lcccc}
  data&config &run&   top-1&top-5\\
  \shline
  \multicolumn{5}{c}{Pretraining on 8 A100s} \\
  IF     &fixed  &102:35 &  \textbf{82.75} & \textbf{96.34} \\
  500\_95& fixed &19:56 &82.64 &96.21 \\
  500\_95& s5  &\baseline{\textbf{17:45}} & \baseline{82.62} & \baseline{96.25}\\
  500\_95&1600 & 39:25 &  82.73&96.31 \\
  \midrule
  \multicolumn{5}{c}{Finetuning on 8 V100s } \\
  IF     & fixed &20:52  & 83.21 & 96.40 \\
  500\_95& fixed &20:55  &83.21 & 96.39\\
  {500\_95}& {s1}    &\baseline{\textbf{18:16}}  & \baseline{\textbf{83.27}} & \baseline{96.35}\\
      \end{tabular}
\end{minipage}
\end{table}

\subsection{Machine Specification}\label{sec:specs}
In our paper, we conduct experiments in three platforms. The specifications are listed in Table~\ref{tab:specs}. We adopt OS cache to enable a memory map, thus, the file system would not be the bottleneck. 

\begin{table}[tbh]

\caption{\textbf{Machine Specifications} of used platforms. For file system, we report the reading speed through \textit{dd} command.}\label{tab:specs}
\tablestyle{2pt}{1.0}
\begin{tabular}{l llll l ll l}
\textbf{Platform} & CPU        & Socks & Cores & Frequency & Memory & GPU   &  & File System \\
\shline
\textit{Jade}             & E5-2698 v4 & 2     & 20    & 2.2 GHz   & 512 GB & V100 SXM2 &         & 2.2 GB/s   \\
\textit{Eureka}           & EPYC 7513  & 2     & 32    & 2.6 GHz   & 512 GB & A100 PCIe &       & 998 MB/s    \\
\end{tabular}
\end{table}

\section{Removing the Data Loading Bottleneck.}

In this section, we present benchmarks on the \textit{eureka} platform (see Appendix~\ref{sec:specs} for details). We investigate the trade-off between image compression and performance. While compressing images by reducing the quality or size can improve the loading throughput, it also comes at the cost of reduced image fidelity and potentially lower overall performance.

We propose a simple yet effective \textit{crop decode} strategy that achieves a significant decompressed data acceleration and a memory usage reduction without compromising the image fidelity. Our enhanced version of FFCV~\cite{leclerc_ffcv_2023}, termed ESSL, achieves a 27.6\% improvement in throughput for a quality level of 90 and a maximum resolution of 500. 

We further explore the impact of compression parameters on the throughput, performance, and storage requirements. To facilitate the discussion, we denote the datasets with compression parameters using the notation Train/Val(res\_quality). We conclude that Train(500\_95) strikes a good balance between these factors and is chosen as the default setting for the training set. Val(500\_100) offers the best performance but at the cost of higher storage requirements. Therefore, Val(500\_100) is used only for reporting final results.

\begin{figure}[t]
    \centering
    \includegraphics[width=.95\linewidth]{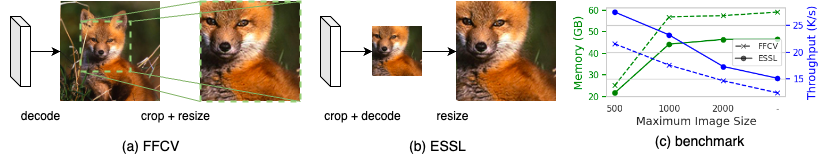}
    \caption{\textbf{Comparison of Throughput} between FFCV and ESSL. `-' denotes no resizing when building the dataset. Maximum image size refers to the largest side (width or height) an image can have after being resized.}
    \label{fig:crop}
\end{figure}

\subsection{Crop Decode}
RandomResizedCrop, a common data augmentation technique prevalent in many machine learning systems, typically involves decoding the entire image before cropping and resizing to the desired size (Fig.~\ref{fig:crop}(a)). However, we propose a novel approach called "crop decode" that decodes only the targeted region, significantly reducing processing overhead (Fig.~\ref{fig:crop}(b)). We term the enhanced FFCV~\cite{leclerc_ffcv_2023}, applying "crop decode", ESSL.
Fig.~\ref{fig:crop}(c) compares data loading performance for compressed datasets using different maximum image sizes with the existing RandomResizedCrop implementation in FFCV (90\% quality). This benchmark involves loading 100,000 images from IN1K, performing random resized cropping to 224x224, and applying random horizontal flip. All benchmarks were executed on the Eureka platform with a batch size of 256, 64 workers, and the OS cache enabled. 
By avoiding unnecessary memory allocation and decoding for discarded regions, our "crop decode" strategy achieves a significant 27.6\% acceleration in data loading, reaching a throughput of 27,493 images/s for a dataset with 90\% quality and a maximum image size of 500. It indicates the potential of ESSL to large-scale datasets with high-quality images.

\subsection{Compression Parameters for Building an FFCV Dataset}

Building an FFCV dataset involves balancing three key factors: throughput, storage usage, and image fidelity. While higher quality or resolution improves image fidelity, it also increases storage requirements and slows down the decoding process. Additionally, storing raw images is impractical for large-scale datasets (e.g., IN1K at 1600 resolution would require 788.97 GB). In such cases, I/O limitations from hardware become the bottleneck for data loading.

Therefore, we focus on compressed datasets and conduct a comprehensive comparison of relevant parameters (resolution and quality). Table~\ref{tab:data_perf} summarizes the file sizes, throughputs, and performance for various compression settings in FFCV. 
It's important to note that datasets undergo two compression processes: one during initial storage as JPEGs and another when building the FFCV dataset. Additionally, JPEG is a lossy compression algorithm, meaning that images may experience some distortion even at the highest quality setting (100).
From the results, we have the following key observations: 1) The highest resolution and quality setting (500\_100) results in the largest file size (181 GB) but does not outperform the default setting in terms of accuracy, which indicates an unknown optimal setting for compression. 2) Lowering the quality to 90 further reduces the file size and increases throughput but at the cost of accuracy. 3) Increasing the resolution to 1000 slightly increases the file size compared to the default setting but reduces the throughput and does not provide a significant accuracy benefit. Therefore, The setting, witha resolution of 500 and a quality of 95, is chosen as the default due to its good trade-off between efficiency (in terms of storage and computational speed) and effectiveness (in terms of model accuracy).

\begin{table}[t]
\centering

\caption{\textbf{Properties of compression parameters} on building FFCV datasets. For each compression setting, we run `finetune' and `pretrain' on the corresponding dataset. The evaluations involve using a MAE-Base/16 model for initializing weights in the `finetune', and for the `pretrain' evaluation, finetuning is performed on the Train(500\_95) dataset. The reported top-1 accuracies are measured on the Val(500\_100). Default settings are marked in \colorbox{baselinecolor}{gray}. }
\label{tab:data_perf} 
\begin{tabular}{llcccc}

\textbf{res} & \textbf{quality} & file size (GB) &throughput (images/s) &finetune & pretrain \\
\shline
\multicolumn{2}{c}{IF} & 138 & 4428 & 83.17 &  82.75 \\
500 &100 & 181 &15796 &83.26  & 82.71\\
500 &95  & \baseline{101} &\baseline{21432} & \baseline{83.22}  & \baseline{82.65}\\
500 &90  & 69 &26079 &83.02  & 82.52\\
1000& 95 & 107 &16476 &83.25  & 82.69\\
256 & 95 & 29 &36056 &82.69  & 81.90\\
\end{tabular}

\end{table}

\subsection{Compression Shift}
Data shift refers to the difference in the statistical properties of two datasets. While the distortions introduced by the compression process are often imperceptible to the human eye, they can be significant enough to impact the performance of machine learning models. Furthermore, different implementations of JPEG compression can also contribute to data shift.

In this study, we compared the compression shift caused by two implementations: PIL\footnote{\href{https://pypi.org/project/pillow/}{Pillow}} applied in IF and jpeg-turbo\footnote{\href{https://libjpeg-turbo.org/}{libjpeg-turbo}} applied in FFCV. To evaluate the impact, we created various validation sets with varying compression qualities and image resolutions. We then assessed the classification performance on these sets using the same fine-tuned weights (ft-MAE-Base/16).

Table~\ref{tab:compression_shift} compares the classification performance on validation sets compressed with PIL and jpeg-turbo. Interestingly, our results indicate that a distortion can even improve the validation accuracy. jpeg-turbo generates a performance drop with same parameters. ft-MAE-Base/16 maintained high performance with a quality of 100\% and a resolution of 500 on the validation set. Therefore, this setting has been chosen as the default for further evaluation.

\begin{table}[tb]
\centering
\caption{\textbf{Classification performance} (top-1) of ft-MAE-B/16 on the IN1K validation set with different compression parameters. The accuracy on the original set is 83.68\%.}
\label{tab:compression_shift}
\subfloat[Image Folder \label{tab:ce_IF}]{
    
\begin{tabular}{y{30}x{25}x{25}x{25}}

 \multirow{2}{*}{\textbf{quality}} & \multicolumn{3}{c}{resolution} \\
 \cmidrule(lr){2-4}
                  & 1000     & 500      & 256      \\
    \shline
100 & 83.69 & 83.71 & 83.58 \\
95  & 83.72 & \textbf{83.73} & 83.11 \\
90  & 83.46 & 83.48 & 82.21 \\
\end{tabular}
}\hspace{5 mm}
\subfloat[FFCV\label{tab:ce_FFCV}]{
\begin{tabular}{y{30}x{25}x{25}x{25}}
{\multirow{2}{*}{\textbf{quality}}} & \multicolumn{3}{c}{resolution} \\
 \cmidrule(lr){2-4}
                  & 1000     & 500      & 256      \\
    \shline
100 & 83.45 & \textbf{83.50} &  83.27\\
95  & 83.46 & 83.49 & 82.96 \\
90  & 83.07 & 83.10 &  82.2\\
\end{tabular}}
\end{table}

\paragraph{Three Augmentation for Compression Shift Mitigation.} 
Data augmentation is a commonly used technique to address data shift by artificially increasing the diversity of the training data and improving the model's generalization ability. In this study, we compared two data augmentation strategies: RandAug~\cite{cubuk_randaugment_2020} and Three Augmentation (3 Aug)~\cite{touvron_deit_2022}.
We initialized the model weights using MAE-Base/16 and finetuned it on both Image Folder (IF) and Train(500\_95) datasets. We then evaluated its performance on the Val(500\_100) and IF datasets using top-1 accuracy.

Our results demonstrate a significant accuracy decline (-0.6\%) for the RandAug trained model when encountering a compression shift between training and validation data. Conversely, the 3 Aug strategy exhibits greater resilience to compression shift, experiencing only a minor performance drop.
Furthermore, Table~\ref{tab:aug_speed} highlights that 3 Aug is significantly more efficient in terms of image processing compared to RandAug.
Therefore, we incorporate 3 Aug into our finetuning recipe to mitigate the compression shift with a minimal sacrifice in performance.

\begin{table}[tbh]
    \centering
\begin{minipage}[t]{.65\linewidth}
\caption{\textbf{Performance drop} between FFCV and IF. Models are initialized with MAE-B/16. }
\label{tab:aug}
\centering
    \begin{tabular}{@{}ll|ccc@{}}

    \multicolumn{2}{c}{Finetune} &      \multicolumn{3}{c}{Test}\\ \shline
    \textbf{Train}        & Aug           & Val(500\_100) & IF   & Drop\\
    IF        & RandAug~\cite{cubuk_randaugment_2020}       & 82.82 & 83.42 & 0.6  \\
    IF       & 3 Aug~\cite{touvron_deit_2022}      & 83.17& 83.44 & 0.27\\
    \midrule
    Train(500\_95)& 3 Aug~\cite{touvron_deit_2022}      & 83.21 & 83.06& 0.15 
    \end{tabular}
\end{minipage} %
\hspace{.02\linewidth} %
\begin{minipage}[t]{.3\linewidth}
    \caption{\textbf{Throughput} of RandAug and 3 Aug. Test with one CPU core on \textit{eureka}.}\label{tab:aug_speed}
    \centering
    \begin{tabular}[t]{lr}
        \multicolumn{2}{r}{\textbf{throughput}}\\
        \shline
        RandAug~\cite{cubuk_randaugment_2020} &901\\ 
        3 Aug~\cite{touvron_deit_2022} & \textbf{1356} 
    \end{tabular}
\end{minipage}
\end{table}

\subsection{Discussion}
While FFCV offers efficient data loading, careful consideration of dataset building parameters remains crucial. Our experiments revealed performance degradation due to compressed datasets, highlighting the trade-off between file size, image fidelity, and model performance. Interestingly, we observed that a controlled level of distortion can even enhance classification accuracy. This suggests that choosing the appropriate combination of resolution, quality, and compression algorithm can potentially further optimize performance. We encourage the community to prioritize this balance when building datasets for efficient training without sacrificing performance.

\section{Progressive Training}\label{sec:resizing}

Building upon the success of curriculum learning in efficient training recipes like Fastai~\cite{howard_fastai_2018} and EfficientNetV2~\cite{tan_efficientnetv2_2021}, we introduce progressive training for both finetuning and pretraining of Vision Transformers (ViTs) to further accelerate the training process. Unlike previous works primarily focused on CNNs and classification tasks, our exploration applies progressive training in the context of ViTs. To the best of our knowledge, while prior research investigated dynamic masking ratio for ViTs~\cite{feng_evolved_2023}, studies on dynamic resolution during pretraining remain scarce.

The pretraining could be beneficial from low-resolution images.
Li et al.~\cite{li_inverse_2023} established an "inverse scaling law" for CLIP training, proposing that fewer image/text tokens are needed for training with larger models due to the regularization effect of resolution~\cite{touvron_deit_2022}. This effect essentially prevents large models from overfitting to the fine-grained details present in high-resolution images.

Recognizing the crucial difference between CNNs and ViTs, particularly the inadaptation of spatial invariance for the linear patch embeddings in ViTs, we first revisit the relationship between perceptual ratio, apparent size, and image resolution. We then investigate the impact of maintaining consistency between perceptual ratio and apparent size during resolution changes for finetuning. Our findings suggest that ViTs are capable to handle the distribution transformation in patches and prioritizing the perceptual ratio during resolution scaling holds greater importance.

In contrast to finetuning, where training difficulty is increased progressively through augmentation, we propose the opposite approach for pretraining: gradual difficulty increase achieved by reducing resolution or increasing masking ratio. Interestingly, not only does reducing image resolution significantly enhance training speed, but it also surprisingly leads to improved performance.

\subsection{Perceptual Ratio, Apparent Size, and Resolution}

We consider an object is modeled in the following: the object in a real size is captured in an image with the image size $ H \times H$. We can infer the apparent ratio of the object  by dividing the image size $S=R/H$, where $R \times R$ is the apparent size of the object. In this context, $R \times R$ represents the object's \textbf{apparent size} within the image. 
However, it's important to acknowledge that data augmentation techniques can alter the image size and apparent size of the object. As discussed in~\cite{touvron_fixing_2022}, the apparent size plays an important role of generalization when transferring from the training set to the validation set. Here, we revisit it and explain the relationship between perceptual ratio, apparent size, and resolution. 
The perceptual ratio refers to the average percentage of the image can be receipted by the model. While, the apparent size refers to real size in pixel of the image.

Our discussion is based on RandomResizedCrop (RRC) which is controlled by a scale parameter $\sigma$ such that $\sigma \sim  U([\sigma^2_-,\sigma^2_+])$ and an aspect ratio $\alpha$ such that $\ln \alpha \sim U([\ln \alpha_-,\ln \alpha_+])$. 
RRC first crops a random region with a random scale and aspect ratio (assuming square images for simplicity, $\alpha=1$). The perceptual ratio of the object, which reflects the most semantic information, is reduced to $\sigma$ after cropping. However, the apparent size, representing the object's perceived size in the image, maintains unchanged. Finally, resizing the cropped region to the output size (h) further scales the apparent size by $ \frac{h}{\sigma H}$:
\begin{equation}
    R_{\mathtt{apparent}} = \mathbb{E}[R \frac{h}{\sigma H}] = \mathbb{E}[C h/\sigma] = C h \frac{3}{2} \frac{\sigma_+^2-\sigma_-^2}{\sigma_+^3-\sigma_-^3},
\end{equation}
where $C=R/H$ is a constant ($C=1$ for simplicity).
The resizing will not change the proportion of the object, therefore, the expectation of perceptual ratio is
\begin{equation}
    R_{\mathtt{ratio}} = \mathbb{E}[\sigma] = \frac{2}{3} \frac{\sigma_+^3-\sigma_-^3}{\sigma_+^2-\sigma_-^2}.
\end{equation}

\begin{figure}[t]
    \centering
    \includegraphics[width=.45\linewidth]{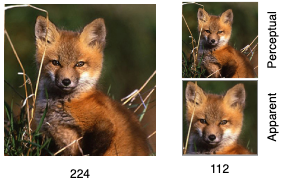}
    \caption{Example of \textbf{perceptual ratio and apparent size}. The perceptual ratio denotes the seen percentage of objects. The apparent size denotes the pixel size for objects.}
    \label{fig:eg_consistency}
\end{figure}

As illustrated in Fig.~\ref{fig:eg_consistency}, the perceptual ratio preserves the semantic information of the fox better when generating low-resolution images, while the apparent size maintains the input distribution for patches. Low-resolution images with the same apparent size can be regarded as local crops~\cite{caron_emerging_2021} which significantly promotes the performance of contrastive learning. Understanding these distinctions is crucial for optimizing training efficiency and performance in ViTs.

\begin{table}[t]

\begin{minipage}[t]{0.47\linewidth}{\begin{center}
\caption{\textbf{Dynamic resizing schemes} for finetuning. $\mathbb{E}[\sigma] $ and $\mathbb{E}[h/\sigma] $ denote the perceptual ratio and apparent size respectively. }
    \label{tab:dres_scheme}
    
    \begin{tabular}{lcccccc}
stage & res  &aug& $\sigma_-^i$ & $\sigma_+^i$   &$\mathbb{E}[\sigma] $&$\mathbb{E}[h/\sigma] $\\ 
\shline

\multicolumn{3}{l}{\textbf{scheme 1}} &&&&\\

1     & 160       &3 Aug& 0.28& 1              &0.47&151\\
2     & 192       &3 Aug&              0.28& 1              &0.47&181\\
3     & 224       &3 Aug+&              0.28& 1           &0.47&211\\

\midrule
\multicolumn{3}{l}{\textbf{scheme 1-}} &&&&\\

1     & 160       &3 Aug& 0.28& 1 &0.47&151\\
2     & 192       &3 Aug& 0.28& 1 &0.47&181\\
3     & 224       &3 Aug& 0.28& 1 &0.47&211\\

\midrule
\multicolumn{3}{l}{\textbf{scheme 1+}} &&&&\\ 

1     & 160       &3 Aug+& 0.28& 1 &0.47&151\\
2     & 192       &3 Aug+& 0.28& 1 &0.47&181\\
3     & 224       &3 Aug+& 0.28& 1 &0.47&211\\
\midrule

\multicolumn{3}{l}{\textbf{scheme 3}} &&&& \\

1     & 160       &3 Aug& 0.28& 0.68 &0.34&210\\
2     & 192       &3 Aug&              0.28& 0.84 &0.41&211\\
3     & 224       &3 Aug+&              0.28& 1           &0.47&211\\

 \end{tabular}
 \end{center}}
\end{minipage} \hspace{3mm}%
\begin{minipage}[t]{.5\linewidth}
        \centering
    \caption{\textbf{Finetuning performance} at the end of each stage for various progressive schemes. The runtime is measured on \textit{Jade}.}
    \label{tab:ft_dres}
    \begin{tabular}{llccc}
 \textbf{scheme} && \multicolumn{3}{c}{Top-1 }\\
 &runtime & 30& 60&100\\
    \shline
  224 & 20:55 &  \textbf{78.93}&  \textbf{81.87}& \textbf{83.21}\\
  192 & 14:51 &  78.64&  81.54& 82.64\\ 
  160 & 8:36&  77.41&  80.60& 82.17\\
  \midrule
  s1 &  18:16   &  \baseline{78.17}&  \baseline{81.73}& \baseline{\textbf{83.27}}\\
  s1- & 18:14   & \textbf{78.74}& \textbf{81.78} &82.78\\ 
  s1+ & 18:20   & 71.61&   77.58 & 82.37 \\
  s3 &  17:39   &  78.51&  81.14& 83.13\\ 
  \\
  \multicolumn{5}{c}{ Converge curves }
    \end{tabular}
    \includegraphics[width=1\linewidth]{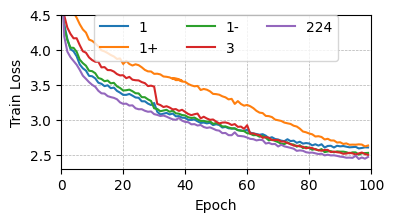}

\end{minipage}
\end{table}

\subsection{Finetuning with Dynamic Resolution Scaling}

In this part, we finetune a pretrained MAE-Base/16 model on Train(500\_95) and evaluate its performance on Val(500\_100). The finetuning recipes are based on the recipe in Table~\ref{tab:recipe}. We implemented dynamic resizing schemes involving three stages, lasting 30, 30, 40 epochs respectively. Table~\ref{tab:dres_scheme} summarizes the details of dynamic resizing schemes.

\paragraph{Perceptual Ratio.} 
\textbf{Default Scheme (Scheme 1)} prioritizes maintaining the perceptual ratio while gradually increasing the image resolution from 160 to 224. Additionally, it strengthens data augmentation in the final stage. Schemes 1, 1-, and 1+ explore the impact of different augmentation strategies. Notably, ``3 Aug+'' denotes a combination of Three Augmentations~\cite{touvron_deit_2022} and ColorJitter (0.3).

\paragraph{Apparent Size.}
\textbf{Scheme 3} focuses on maintaining the apparent size throughout training while progressively raising the perceptual ratio, mimicking a shift from local to global cropping.

Table~\ref{tab:ft_dres} compares the finetuning performance under various progressive schemes. The results reveal that the model is struggle to learn with strong augmentation applied in the initial stage. Gradual augmentation enhancement leads to a better generalization.

\subsection{Pretraining with Dynamic Resolution Scaling}
While data augmentation proves beneficial for finetuning in progressive training, research suggests it can be harmful during pre-training with Masked Image Modeling (MIM)~\cite{he_masked_2022,cao_rethinking_2021}. Therefore, we explore two alternative strategies to increase the difficulty during pretraining: masking ratio and masking size.
Table~\ref{tab:predres_scheme} details our pretraining schemes.

We apply online prob to monitor the quality of learned representations during pretraining, where a linear classifier is attached to the frozen features. 
To assess the performance of different pretrained models, we finetune them on the Train(500\_95) and evaluate their Top-1 accuracy on the Val(500\_100) using a fixed resolution of 224.

\begin{table}[t]

\begin{minipage}[t]{0.45\linewidth}
    \centering
    \caption{Dynamic resizing scheme for pretraining. $m$ is the masking ratio. }
    \label{tab:predres_scheme}
    \begin{tabular}{lccc|ccc}
    \textbf{Stage} & res. &$m$  && res. &$m$  &\\ \hline
     & 1&  &&  4& &\\
    
    1     & 160      &0.50  && 160      &0.75  &\\
    2     & 192      &0.66  && 192      &0.80  &\\
    3     & 224      &0.75  && 224      &0.85  &\\
    \midrule
     & 2&  &&  3& &\\
    
    1     & 160      &0.75  && 224      &0.75  &\\
    2     & 192      &0.75  && 192      &0.75  &\\
    3     & 224      &0.75  && 160      &0.75  &\\
    \midrule
    & 5&  &&  6& &\\
    1     & 224      &0.75  && 224      &0.75  &\\
    2     & 192      &0.75  && 192      &0.75  &\\
    3     & 224      &0.75  && 224      &0.85 &\\
    
     \end{tabular}
\end{minipage} 
\hspace{5mm}%
\begin{minipage}[t]{.5\linewidth}
        \centering
    \caption{ \textbf{Pretraining performance} for different progressive schemes. We run experiments on \textit{Eureka}.} \label{tab:premae_dres}
    \begin{tabular}{llccc}
 \multirow{2}{*}{\textbf{Scheme}}&  \multirow{2}{*}{Time} &\multirow{2}{*}{Online prob}& \multicolumn{2}{c}{Finetune}\\
     &  && Top-1& Top-5\\    
    \shline
\multicolumn{4}{c}{\textbf{Fixed Size}}           &                \\    
224  & 19:56 & 62.96          & 82.64          & 96.21          \\
192  & 08:39 & -              & 82.48          & 96.16          \\
160  & 08:07 & -              & 82.24          & 96.12          \\
\midrule
\multicolumn{4}{c}{\textbf{Dynamic Resizing}}           &                \\
s1    & 17:57 & 60.79          & 82.31          & 96.06          \\
s2    & 16:46 & 63.66          & 82.46          & 96.21          \\
s3    & 14:39 & 63.77 & 82.53          & 96.17          \\
s4    & 15:39 & \textbf{64.51} & 82.49 & 96.15 \\
s5    & 17:45 & \baseline{62.96}          & \baseline{\textbf{82.62}} & \baseline{\textbf{96.25}} \\
s6    & 17:37 & 62.86          & 82.41          & 96.15          \\
    \end{tabular}
\end{minipage}
\end{table}

\paragraph{Masking Ratio.}
The percentage of the masked inputs in MIM~\cite{atito_gmml_2022,he_masked_2022}, masking ratio,  
 plays a crucial role by controlling how much information the model drops during training. A higher ratio signifies a greater difficulty in inferring image semantics.
To investigate the impact of masking ratio in progressive pre-training, we designed two schemes (Table~\ref{tab:predres_scheme}) as well as the baseline (\textbf{Scheme 2}).  \textbf{Scheme 1} considers the difficulty increase due to both reduced image size (lower resolution) and masked parts. Therefore, it starts with a lower masking ratio in the initial stage. Conversely, \textbf{Scheme 4} directly increases the masking ratio from 0.75 to 0.85, significantly reducing training time. 
It is notable that the pretraining performance is consistently improved by increasing the masking ratio, especially the online prob performance. 
This indicates a potential of improvements by tuning the masking ratio. As a result, scheme 4 saves 6.6\% training time with 0.75\% accuracy improvement for online prob. 

\paragraph{Masked Perceptual Ratio.}
In Masked Image Modeling (MIM), where the prediction unit is a patch, we define the masked perceptual ratio as the average proportion of a patch masked during training. This ratio is calculated as the expected value of the product of the masking ratio ($\sigma$) and the patch size (p) divided by the output size (h):
\begin{equation}
    R^m_{\mathtt{ratio}} = \mathbb{E}[\sigma \frac{p}{h}].
\end{equation}

Intuitively, reducing the image size increases the masked area within each patch, as a larger portion of the object is represented by each smaller patch. This leads to a higher masked perceptual ratio. 

We designed \textbf{Schemes 3} that progressively change the image size in opposite directions. Compared to Scheme 2, Scheme 3 achieves slightly better performance and saves roughly 12.62\% training time.

\paragraph{Additional Schemes.}
\textbf{Scheme 5} was proposed to address information loss in low-resolution images. \textbf{Scheme 6} is a hyber design of increasing the masking ratio and the masked perceptual ratio during training. 

From the results, we can see that Scheme 5 achieves the best finetuning performance and Scheme 4 has the highest online prob performance. However, Scheme 6 shows poor performance on both two evaluations. In conclusion, training models on lower resolutions (higher masked perceptual ratios) can enhance capabilities, but with limitations due to missing information. Finding the right balance between leveraging lower resolutions and mitigating information loss is crucial.

\paragraph{Image Folder.}
We further investigated the effectiveness of progressive training without FFCV. The models were pretrained on IF. Similar to the findings with the FFCV dataset, the results in Table~\ref{tab:dres_if} reveal a consistent trend. Scheme 4, which directly increases the masking ratio during pre-training, once again exhibits the highest improvement in online probing performance. Additionally, it achieves an even greater training time reduction of 18.5\% compared to the fixed-size scheme.

\begin{table}[h]
    \centering    
    \caption{Progressive training without FFCV. We run pretraining on \textit{Jade}.}
    \label{tab:dres_if}
\begin{tabular}{lrccc}
 \multirow{2}{*}{\textbf{Scheme}}&  \multirow{2}{*}{Time} &\multirow{2}{*}{Online prob}& \multicolumn{2}{c}{Finetune}\\
     &  && Top-1& Top-5\\    
    \shline               
  224 & 102:35 & 60.98 & \textbf{82.75} & \textbf{96.34} \\
   s2 & 87:01 & 61.85& 82.66& 96.28 \\
   s4 & \textbf{83:35} & \textbf{62.5} & 82.64& 96.28 \\
    \end{tabular}
\end{table}

\section{Conclusion}

\paragraph{Limitation.} 
Our current machine learning library is limited to masked autoencoders. We plan to expand its capabilities to encompass a wider range of fundamental architectures in the future. Additionally, while our library utilizes lossy compression algorithms, these methods can introduce performance degradation in SSL models. Our findings highlight the need of compression algorithms for preserving performance.

Despite these limitations, this work demonstrates the potential of dynamic resolution training for accelerating training and potentially improving model performance. Our library serves as a valuable tool for rapid algorithm validation, facilitating future studies on SSL. The discussion about compression parameters could help researchers to build proper datasets with regard to their hardware systems.

%
%
\bibliographystyle{splncs04}
\bibliography{references}

\newpage
\appendix

\section{Training Benchmark}
\begin{figure}[tbh]
\centering

\includegraphics[width=.8\linewidth]{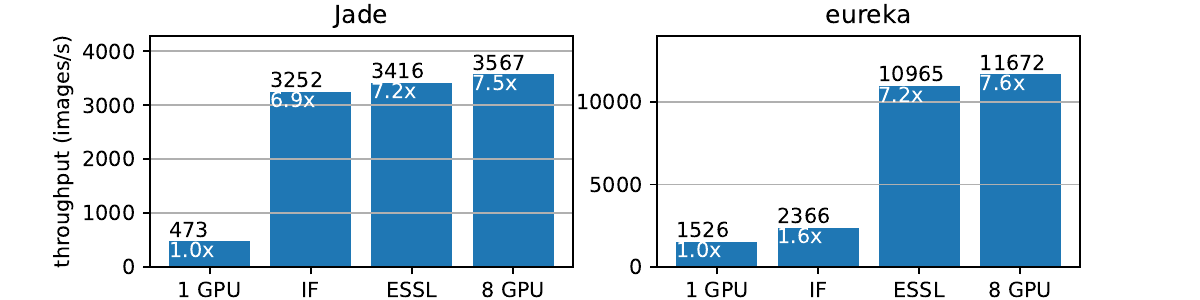}
\caption{\textbf{Training benchmark} of throughput (images/s) on two platforms, including data loading, forward, backward, and optimization. GPU 1 and 8 denote the pure training processes without data loading. }\label{fig:train_perf}
\end{figure}

We benchmark the training throughput of Image Folder and FFCV on two platforms (Jade and eureka). Each step includes data loading, forward, backward, and optimization. To demonstrate the ideal speed, we exclude the data loading and employ the result trained with 1 GPU as the reference. The experiments are conducted with a batch size of 256 and 10 workers per process. As shown in Figure~\ref{fig:train_perf}, ESSL reaches a speed up ratio of 7.2x on both Jade and on eureka, which is very close to the ideal situation (GPU 8). We observe that the throughput on eureka is smaller than on Jade even through better GPUs are equiped. This is caused by the IO bound of reading images from a network-based file system when the cluster is busy.

\section{Pipeline Decomposition}

We decompose the pipeline in Table~\ref{tab:decomp} for loading a FFCV dataset with resolution of 500 and quality of 90. The Simple augmentation consists of RandomHorizontalFlip, converting to pytorch format, and normalization. The experiments are run on eureka.
\begin{table}[htb]
    \centering    
    \caption{\textbf{Pipeline Decomposition} of FFCV and ESSL. Our \textit{crop decode} strategy accelerates the decoding process.}
    \label{tab:decomp}
    \begin{tabular}{rlcrl}
 \textbf{{throughput}}&   && &\\
 154895.22&  \quad read only && &\\
 \shline
 \multicolumn{2}{c}{FFCV}&& \multicolumn{2}{c}{ESSL}\\
 
 33556.75& +decode      && {42403.61}&+crop decode\\
 21448.03& +crop resize && {26825.82}&+resize\\
 11665.95&  +Simple     && {15439.60}&+Simple\\
 \end{tabular}
\end{table}

\section{Data Visualization}

\begin{figure}
    \centering
    
    \subfloat[Image Folder]{\includegraphics[width=0.32\linewidth]{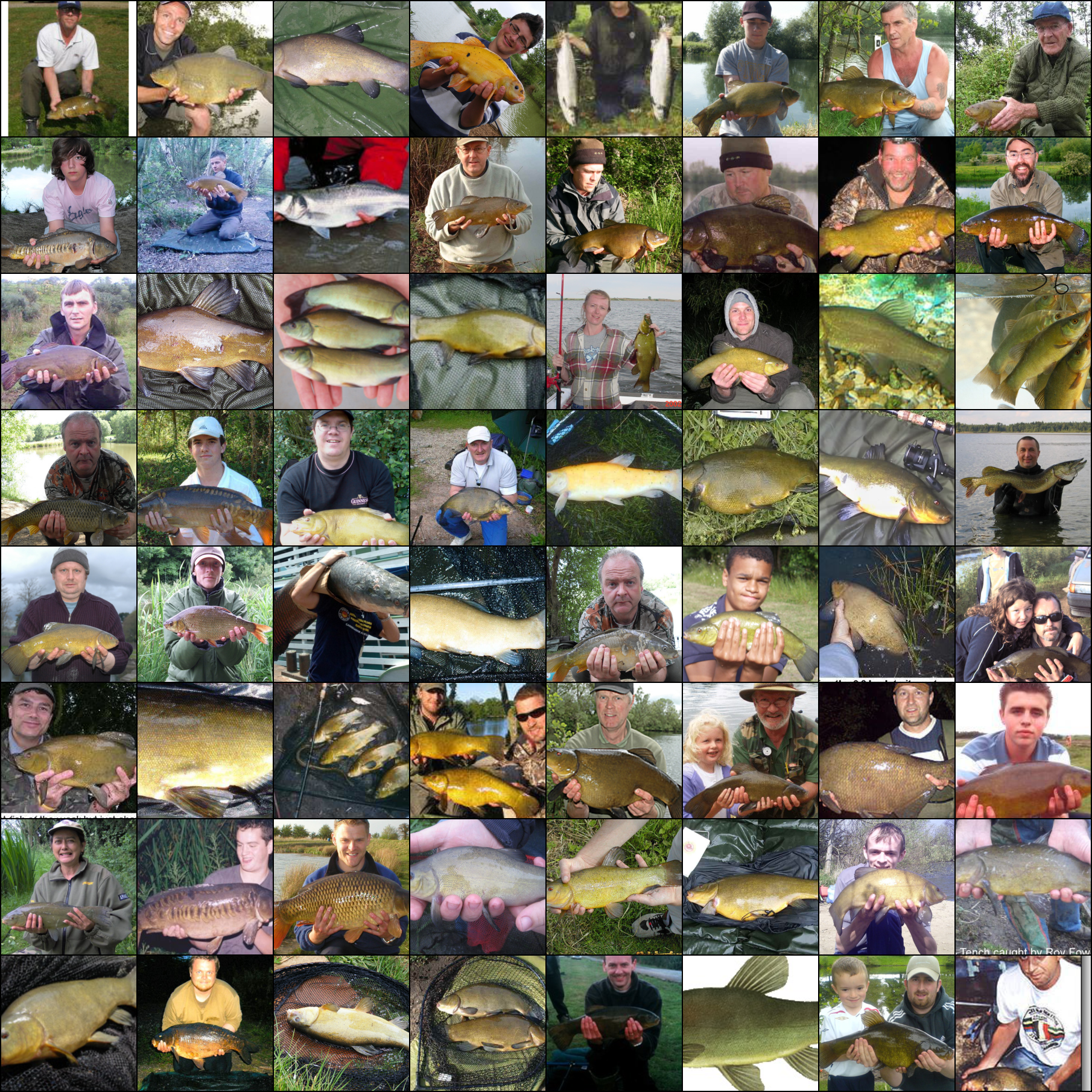} } %
    \subfloat[FFCV]{\includegraphics[width=0.32\linewidth]{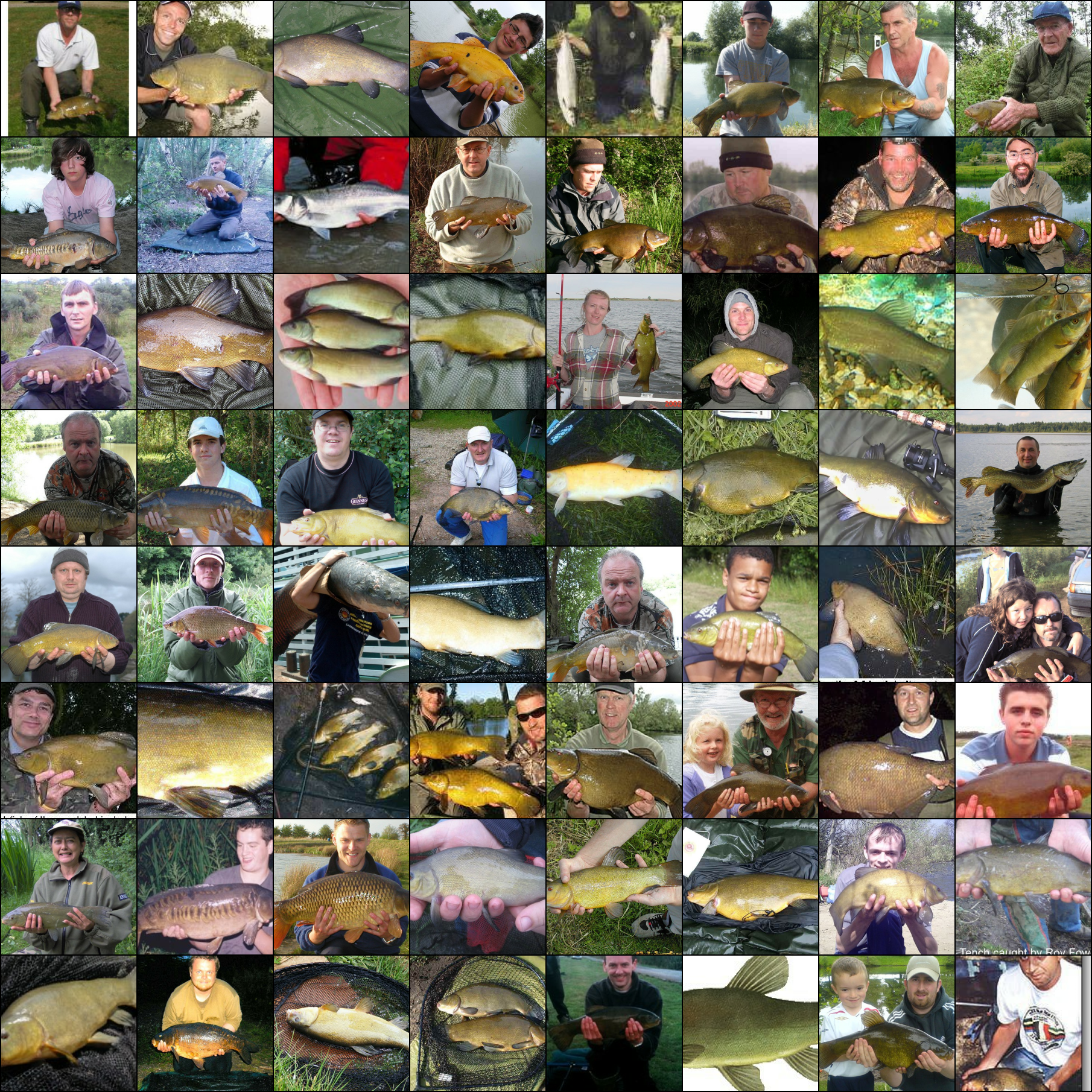} }%
    \subfloat[Difference]{\includegraphics[width=0.32\linewidth]{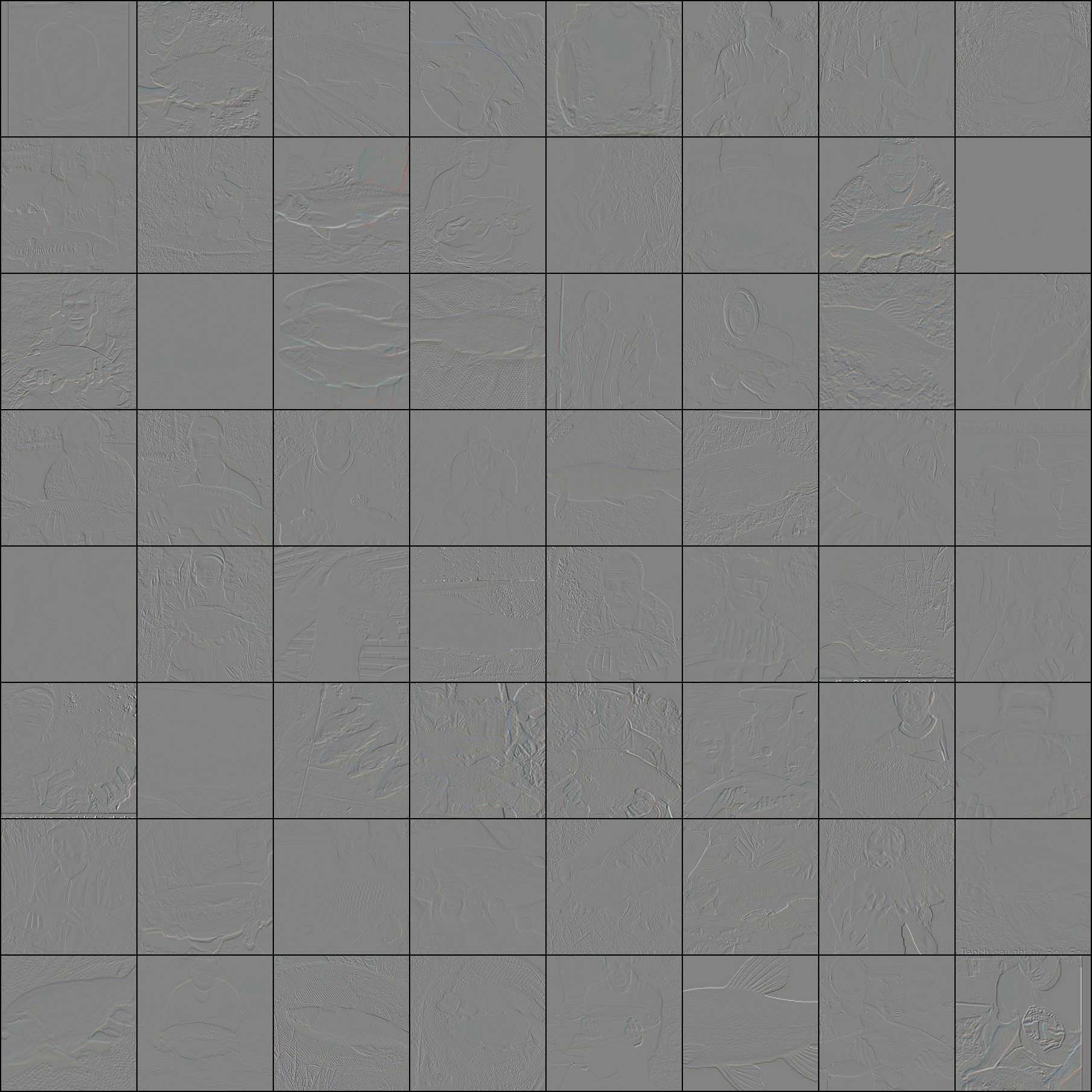} }
    \caption{Visualization of samples from Image Folder and FFCV, and the difference between them.}
    \label{fig:viz_sample}
\end{figure}

We visualized samples from Image Folder and FFCV in Figure~\ref{fig:viz_sample}. The samples are the first 64 images in the train set of IN1K. The difference denotes the errors between the samples from two data loaders. The average absolute error is 0.025. We could observe outlines of objects for the differences, which means some key information about objects is omitted.

\end{document}